\documentclass[10pt,conference]{IEEEtran}
\IEEEoverridecommandlockouts
\usepackage{cite}
\usepackage{threeparttable}
\usepackage{amsmath,amssymb,amsfonts}
\usepackage{enumitem}
\usepackage{multirow}
 \usepackage{bigdelim}
\usepackage{array} 
\usepackage[super]{nth}
\usepackage[table,x11names]{xcolor}
\usepackage{float}
\usepackage{subcaption} 
\usepackage{siunitx}
\usepackage{tikz} 
\usepackage{subcaption} 
\usepackage{algorithmicx}
\usepackage{algpseudocode}
\usepackage{algorithm}
\usepackage{algpseudocode}
\usepackage{graphicx}
\usepackage{textcomp}
\usepackage{pifont}
\usepackage{makecell}
\usepackage{booktabs}
\usepackage{todonotes}
\usepackage{flushend} 
\usepackage[hidelinks]{hyperref}
\usepackage{algorithm}
\usepackage{algorithmicx}
\usepackage{algpseudocode}
\usepackage{float}
\usepackage{booktabs} 
\usepackage{makecell}
\usepackage{overpic}

\usepackage[T1]{fontenc}
\usepackage[a-1b]{pdfx}

\newcommand{\subfiglabel}[1]{\put(73,70){\color{white}\footnotesize\textbf{(#1)}}} 

\definecolor{softgreen}{RGB}{34, 139, 34} 

\author{
     \IEEEauthorblockN{
         Theofanis Vergos\IEEEauthorrefmark{1},
         Polykarpos Vergos\IEEEauthorrefmark{1},
         Mehdi B. Tahoori\IEEEauthorrefmark{1},
         Georgios Zervakis\IEEEauthorrefmark{4}
     }
     \IEEEauthorblockA{
        \IEEEauthorrefmark{1}Karlsruhe Institute of Technology, Germany,
         \IEEEauthorrefmark{4}University of Patras, Greece
     }
     \IEEEauthorblockA{
        \IEEEauthorrefmark{1}\{theofanis.vergos, polykarpos.vergos, mehdi.tahoori\}@kit.edu,
         \IEEEauthorrefmark{4}zervakis@ceid.upatras.gr
     }
}

\begin{document}
\bstctlcite{IEEEexample:BSTcontrol}

\title{
\vspace{-0.9cm}
{\small This article is accepted for publication in \textit{IEEE Design, Automation \& Test in Europe} (DATE) 2026}\\[-1.8ex]
\rule{\textwidth}{0.4pt}
Bespoke Co-processor for Energy-Efficient Health Monitoring on RISC-V-based Flexible Wearables
\vspace{-0.3cm}
}

\maketitle

\begin{abstract}
Flexible electronics offer unique advantages for conformable, lightweight, and disposable healthcare wearables.
However, their limited gate count, large feature sizes, and high static power consumption make on-body machine learning classification highly challenging.
While existing bendable \mbox{RISC-V} systems provide compact solutions, they lack the energy efficiency required.
We present a mechanically flexible \mbox{RISC-V} that integrates a bespoke multiply-accumulate co-processor with fixed coefficients to maximize energy efficiency and minimize latency.
Our approach formulates a constrained programming problem to jointly determine co-processor constants and optimally map Multi-Layer Perceptron (MLP) inference operations, enabling compact, model-specific hardware by leveraging the low fabrication and non-recurring engineering costs of flexible technologies. Post-layout results demonstrate near-real-time performance across several healthcare datasets, with our circuits operating within the power budget of existing flexible batteries and occupying only $2.42$mm$^2$, offering a promising path toward accessible, sustainable, and conformable healthcare wearables. Our microprocessors achieve an average $2.35$x speedup and $2.15$x lower energy consumption compared to the state of the art.
\end{abstract}

\vspace{-0.5cm}

\section{Introduction}
\setlength{\parskip}{0pt}  
\label{sec:introduction}
The growing emphasis on personalized and preventive healthcare has intensified the need for continuous monitoring of physiological signals outside clinical settings~\cite{khan2024anomaly}.
Wearable devices play a central role in this shift, enabling tracking of vital signs during daily activities~\cite{mahato2024hybrid,jeong2025exploiting}.
Although health monitoring could greatly benefit from---and, in many cases, outright necessitates---flexible, conformable, and disposable devices~\cite{mishra:2020:commodityHardware,Shatta:ICCAD2025}, most research has focused on Machine Learning (ML), a key enabler for such systems that extracts insights from complex biosignals~\cite{Kumar2021HierarchicalDN,Aqajari2020GSRAF,Shiyi:IoT2022:StressMonitoring,B-HAR:pravadelli, CNN:Margaria,Fog:pravadelli}, while flexible hardware remains overlooked~\cite{Shatta:ICCAD2025}.
Most commercial wearables rely on non-disposable, rigid, silicon systems, hindering comfort and skin conformability, while combined with their relatively high costs limit affordability and ubiquitous adoption~\cite{mishra:2020:commodityHardware,Shatta:ICCAD2025}.


Flexible electronics have emerged as a promising solution, offering ultra-thin, lightweight, and stretchable circuits ideal for conformable health-monitoring systems~\cite{Flexible:Devenuto}.
Built on flexible substrates, flexible integrated circuits naturally conform to body contours, enhancing comfort during extended wear~\cite{Gao:Nature2022:FlexibleSensor}.
In this work, we target the FlexIC process of Pragmatic Semiconductor~\cite{flexic_gen3}, which uses Indium Gallium Zinc Oxide (IGZO) thin-film transistors (TFTs).
FlexIC enables rapid production cycles and extremely low-cost, even portable, fabrication~\cite{flexicores,ozer:nature2024:bendableRiscV} of disposable hardware well-suited for inexpensive patches, valuable for both commercial applications and clinical settings.
Moreover, its low-temperature fabrication process promotes sustainable manufacturing, achieving orders-of-magnitude reductions in carbon footprint and water consumption compared to traditional silicon technologies~\cite{ozer:nature2024:bendableRiscV}.

However, flexible electronics face fundamental limitations, including low integration density, large feature sizes, and limited performance~\cite{tahoori2025computing}.
For example, current FlexICs integrate up to $20$K NAND2-equivalent gates~\cite{tahoori2025computing} and operate at frequencies of a few hundred kHz~\cite{flexic_gen3}.
Additionally, the absence of p-type devices results in resistor-nMOS logic and elevated power consumption.
These constraints make the integration of complex circuits---such as the ML classifiers typically required in healthcare wearables---extremely challenging.

Flexible microprocessors have been explored in~\cite{biggs2021natively,flexicores,raisiardali2025flexingriscvinstructionsubset}, but they are either too large~\cite{biggs2021natively} or lack the computational power to support ML tasks~\cite{flexicores,raisiardali2025flexingriscvinstructionsubset}.
Ozer et al.~\cite{ozer:nature2024:bendableRiscV} presented and fabricated Flex-RV, the first Bendable RISC-V microprocessor.
Focusing on area efficiency, Flex-RV builds on the serial RISC-V variant, SERV~\cite{kindgren2019serv}, and integrates a multiply-accumulate (MAC) co-processor to improve system latency for neural network inference.
The co-processor in the \mbox{Flex-RV} relies on conventional arithmetic units, overlooking the hardware efficiency of bespoke implementations. 
Similarly, the latency and energy efficiency of Flex-RV are limited by the need to transfer both activations and weights to its co-processor due to its conventional implementation.

Flexible electronics and bespoke design mutually enable each other.
On one hand, the extremely low fabrication costs of flexible electronics, combined with their short production timelines and low non-recurring engineering costs~\cite{tahoori2025computing}, enable the design of bespoke (i.e., fully customized) circuits tailored to the target applications, e.g., ML circuits with hardwired coefficients.
On the other hand, the unmatched hardware efficiency of bespoke designs~\cite{Ozer:Nature:2020, Ozer:DFLEPS2020,velu24,Mubarik:MICRO:2020:printedml,Armeniakos:DATE2022:axml,Armeniakos:TC2023:codesign,Kokkinis:TC:2024:enabling,Mrazerk:ICCAD2024,Afentaki:ICCAD23:hollistic,Afentaki:DATE2024:embedding} can help overcome inherent technology limitations, enabling realistic designs and unlocking a wide range of applications for FlexICs, such as ML-based healthcare monitoring.

In this work, we overcome these limitations by enabling bespoke ML acceleration within flexible RISC-V–based systems. We integrate the compact SERV core with a specialized co-processor that, with minimal area overhead, maximizes energy efficiency and minimizes inference latency for MAC-intensive classifiers such as Multi-Layer Perceptrons (MLPs).
These design choices are particularly important for flexible healthcare wearables, which are battery-powered or rely on energy harvesters, thus demanding exceptional energy efficiency, low power, and compact circuitry to fit within small, lightweight form factors. Meanwhile, MLPs have been shown to achieve state-of-the-art accuracy in healthcare monitoring applications~\cite{afentaki2025islped}, making our solution well-suited for this domain. Exploiting the low fabrication cost, short lifecycle, and disposable nature of flexible circuits, we tailor the microprocessor design to each target MLP. Aligning the rapid FlexIC production timelines with equally fast design cycles, we propose an automated framework that uses constrained programming and (i) derives the fixed coefficients of the bespoke co-processor for each MLP and (ii) optimally maps MLP inference operations onto the hardware by decomposing each neuron’s products across the available bespoke multipliers.

Post-layout evaluation across healthcare datasets demonstrates that our circuits achieve $<\!1$s inference latency, i.e., near-real-time operation~\cite{afentaki2025islped}, with an energy consumption of only $0.48$mJ per inference, while meeting conformability and stretchability requirements.
Compared to the state-of-the-art \mbox{Flex-RV}~\cite{ozer:nature2024:bendableRiscV}, we achieve $2.35$x lower latency and $2.15$x lower energy consumption.
\textbf{Our novel contributions are as follows:}
\begin{enumerate}[topsep=0pt,leftmargin=*]
\item To the best of our knowledge, we propose and design the first bespoke accelerated RISC-V microprocessor for flexible ML-based applications.
\item We propose an automated methodology to jointly select the constant parameters of the bespoke co-processor and determine the optimal MLP to co-processor mappings, thereby minimizing inference latency and energy.
\end{enumerate}

\section{Background}

Pragmatic’s Gen3 FlexIC~\cite{flexic_gen3} platform enables the fabrication of flexible integrated circuits on $200$mm/$300$mm polyimide wafers.
FlexIC employs n-type metal-oxide IGZO TFTs, achieving a minimum channel length of $0.6$\si{\micro\meter}.
The polyimide base features a thickness of less than $30$\si{\micro\meter}, allowing for ultra-thin circuits.
FlexICs have demonstrated mechanical robustness and can bend without any damage to circuitry~\cite{tahoori2025computing}.

IGZO TFTs combine mechanical flexibility with compatibility for low-temperature, cost-efficient fabrication~\cite{ozer:nature2024:bendableRiscV}.
By integrating all components directly on the substrate, the process avoids rigid wafers, high-temperature steps, and additional packaging, producing lightweight and conformable devices that are robust under bending and stretching.
The streamlined manufacturing significantly reduces prototype development time, allowing lab-scale devices to be fabricated in days rather than months, and typically delivers finished circuits in about six weeks.
The intrinsic flexibility further eliminates the need for encapsulation layers, simplifying production and improving reliability under mechanical stress.

\begin{figure}[t!]
    \makebox[\linewidth][l]{\hspace{0.5cm}%
        \includegraphics[width=\linewidth]{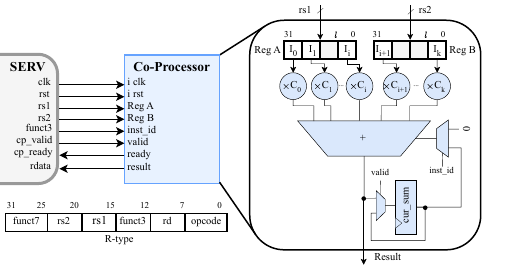}}
    \caption{Our flexible RISC-V-based systems comprising a SERV core and a bespoke co-processor.}
    \label{fig:serv_coprocessor_interface}
    \vspace{-4ex}
\end{figure}

\section{Proposed Bespoke Accelerated RISC-V}
\label{sec:bespoke}

\subsection{System Overview}
Our RISC-V based system for flexible ML-powered wearables is illustrated in Fig.~\ref{fig:serv_coprocessor_interface}.
Specifically, we use the SERV core tightly coupled with our bespoke co-processor.
SERV employs a bit-serial arithmetic logic unit (ALU), shift registers to minimize hardware complexity, and a simple finite-state machine (FSM) for control.
By processing one bit at a time, SERV achieves exceptionally low area and power consumption, making it well-suited for flexible electronics applications.
Moreover, SERV has already been fabricated with FlexIC and tested under varying tension and compression corners, demonstrating both feasibility and robustness~\cite{ozer:nature2024:bendableRiscV}.
Compared to other fabricated bendable microprocessors~\cite{flexicores}, SERV is powerful enough to support more complex requirements, e.g., as in flexible healthcare monitoring wearables.

The proposed co-processor reuses the operands and result signals of R-type instructions and the decoded \textit{funct3} field to distinguish co-processor operations (see Fig.~\ref{fig:serv_coprocessor_interface}).
Custom instructions, issued via inline assembly, are added to the SERV ISA and seamlessly supported by minor modifications to SERV’s decode unit, which detects them using the opcode and the 25\textsuperscript{th} instruction bit.
Upon detection, the decoder triggers the write-back path and notifies the FSM and related modules.
Data transfer and synchronization follow a ready/valid handshake: SERV issues transactions only when the co-processor is ready, and the pipeline stalls until a valid response is received.

With the co-processor seamlessly integrated, determining its most efficient architecture is crucial.
To maximize hardware efficiency, we propose a bespoke MAC-based co-processor.
Specifically, instead of costly conventional multipliers, we use bespoke multipliers, where one operand is a runtime input and the other is a predefined constant.
However, this design choice requires efficiently folding the entire MLP inference over the limited set of available constants, a challenge that is addressed, together with optimal constant selection, in Section~\ref{sec:optimization}.

Two main benefits are obtained from our proposed design.
First, using bespoke multipliers, hardware savings (area and power) are obtained from their simpler implementation. 
For example, Fig.~\ref{fig:bespoke_mul} shows the area of flexible bespoke signed multipliers for $4$-bit inputs and compares it with the area of a conventional $4$x$4$-bit signed multiplier. 
Note that we consider fixed-point arithmetic in ML inference, allowing all operations to be performed entirely with integer operations.
As expected, the area of the bespoke multipliers varies with the constant value.
Nevertheless, on average, bespoke multipliers require $3$x less area. Secondly, since the coefficients are hardwired into the co-processor, the entire SERV-to-ML co-processor bandwidth (i.e., registers \texttt{rs1} and \texttt{rs2} in each call) can be dedicated solely to transferring inputs (e.g., activations in MLPs).
Leveraging this and trading some of the area savings from the bespoke multipliers (see Fig.~\ref{fig:bespoke_mul}), we can instantiate more multipliers, increasing the number of MAC operations per $64$-bit transfer from SERV to the co-processor, which can reduce inference latency and lower energy consumption.

\vspace{-0.5ex}
\subsection{Bespoke Co-processor}

The proposed co-processor comprises several by-constant multipliers, an adder tree to accumulate the products and add them with the current partial sum, along with two multiplexers for control.
The first, controlled by \texttt{inst\_id} (the \texttt{funct3} field in the RISC-V R-type instruction), selects between 0---to start a new weighted-sum computation---and the current sum for ongoing computations.
The second signal, controlled by \texttt{valid}, ensures that only correct sums, i.e., after \texttt{rs1} and \texttt{rs2} have been fully received, are written to the \texttt{cur\_sum} register.
For implementation simplicity and to avoid the overhead of multiplexers and control circuitry, each bespoke multiplier input is fixed and directly connected to predefined bit positions of the two input registers.
Thus, assuming $l$-bit inputs, if the product $x \times C_0$ is required, $x$ must be placed at position $I_0$ of register \texttt{reg A} (see Fig.~\ref{fig:serv_coprocessor_interface}), i.e., in the $l$ most significant bits (MSBs) of the instruction's \texttt{rs1} register.

\begin{figure}[t!]
    \centering
    \hspace{-1cm} 
    \includegraphics[width=0.8\linewidth]{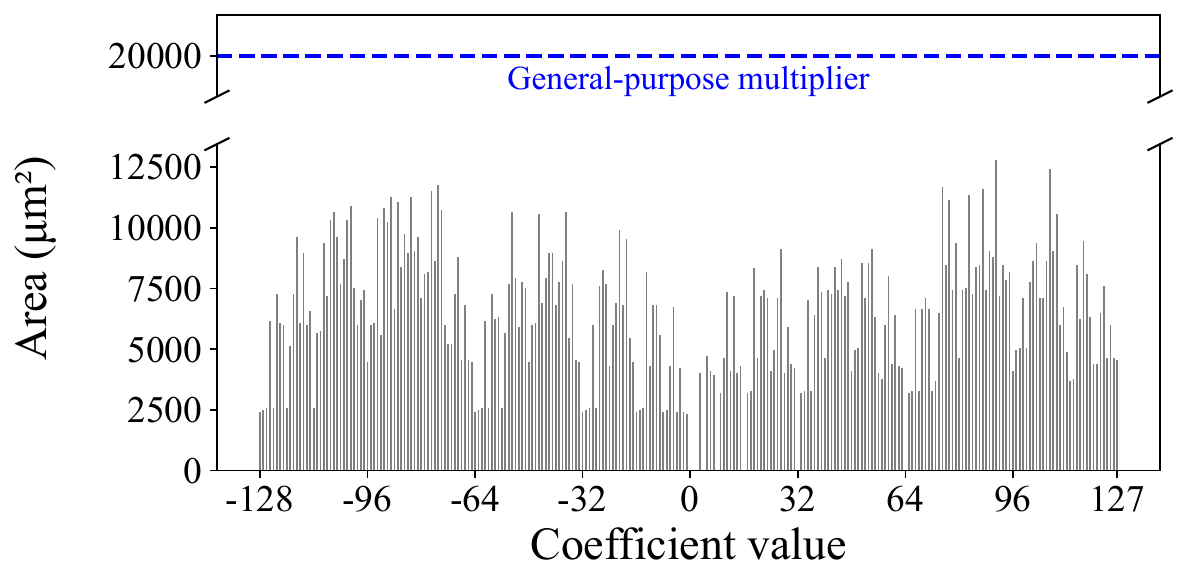}
    \caption{Area of signed bespoke multipliers in FlexIC technology~\cite{flexic_gen3} for varying constants.
    For reference, conventional $4$x$4$-bit and $8$x$4$-bit multipliers occupy $0.020$ mm$^2$ and $0.038$ mm$^2$, respectively. Post-synthesis results are reported.}
    \label{fig:bespoke_mul}
    \vspace{-3ex}
\end{figure}

\begin{figure}[t!]
    \centering 
        \includegraphics[width=\linewidth]{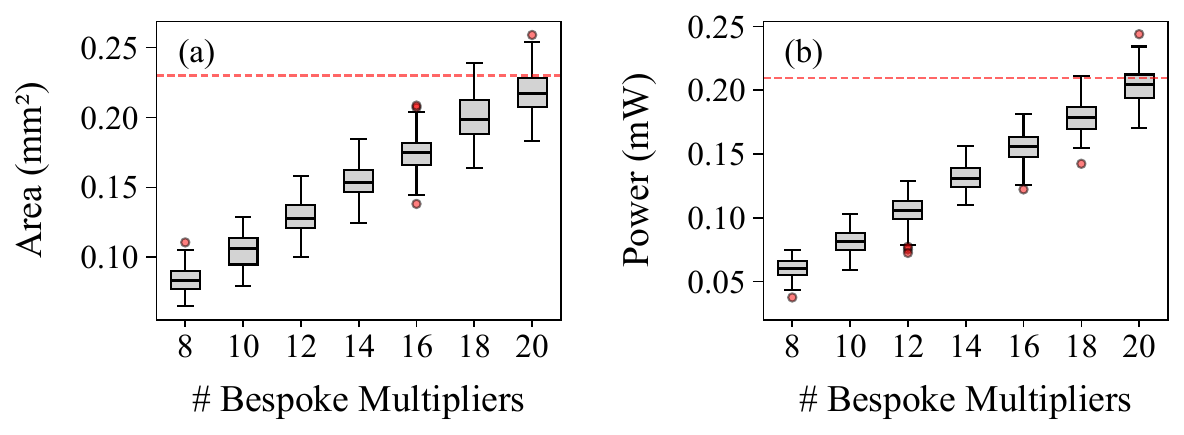}
        \vspace{-2ex}
    \caption{Monte Carlo analysis of (a) area and (b) power of our bespoke co-processor as a function of the number of bespoke multipliers. Results are on $200$ post-synthesis samples with $4$-bit inputs and random constants in $[-8, 7]$. The red line refers to a co-processor with eight $4$x$4$ conventional multipliers.}
    \label{fig:mc}
    \vspace{-3ex}
\end{figure}

Although flexible applications are highly resource-constrained, they typically feature low precision requirements~\cite{Henkel:ICCAD2022:expedition}.
Inputs in the range of $4$ or $5$ bits have been shown sufficient for most flexible ML applications~\cite{tahoori2025computing,Ozer:Nature:2020,Armeniakos:TC2023:codesign}.
As deduced from Fig.~\ref{fig:bespoke_mul}, the area of our co-processor may vary depending on the constant values selected for the bespoke multipliers.
Fig.~\ref{fig:mc} evaluates configurations with $4$-bit inputs and varying number of bespoke multipliers.
Specifically, we perform a Monte Carlo analysis, randomly sampling each bespoke multiplier’s constant.
As shown, area and power increase with the number of bespoke multipliers and vary with the selected constants.
Though, the dominant factor is the total number of multipliers, as evidenced by the minimal overlap among the box plots, primarily due to the growing contribution of the adder tree with more summands.

Fig.~\ref{fig:mc} also shows the area and power of a conventional co-processor that employs eight $4$x$4$-bit conventional multipliers (red line).
Since each conventional multiplier requires $8$ input bits in total ($4$ bits for each operand), only eight such multipliers can be instantiated given that registers \texttt{reg A} (\texttt{rs1}) and \texttt{reg B} (\texttt{rs2}) together provide $64$ bits.
The area and power of this conventional co-processor are $0.23$mm$^2$ and $0.21$mW, respectively.
As shown in Fig.~\ref{fig:mc}, for $4$-bit inputs, up to sixteen bespoke multipliers can be instantiated without incurring any area or power overhead compared to the conventional co-processor with eight $4$x$4$-bit multipliers.
Moreover, since each bespoke multiplier requires only $4$ input bits, sixteen is also the maximum number of bespoke multipliers that can be simultaneously fed by \texttt{reg A} and \texttt{reg B}.
Identical results are obtained for $5$-bit inputs, where twelve bespoke multipliers can be instantiated, compared to a conventional co-processor that integrates seven $5$x$4$-bit multipliers.

Nevertheless, as aforementioned, one drawback of using bespoke multipliers is that some products may need to be decomposed into multiple partial products if the required constant is not directly available.
For example, assume that our co-processor has sixteen constant multipliers, with $C_0 = 3$, $C_1 = 15$, and $C_i \neq 18$, $\forall i > 1$.
If the product $x \times 18$ is required, it can be computed by storing $x$ in both the $I_0$ and $I_1$ positions (i.e., the $8$ MSBs of the instruction's \texttt{rs1} register).
Although this approach produces the correct result, two bespoke multipliers are used to compute a single product.

The key takeaway from Fig.~\ref{fig:mc} is that, while the selected constant values do influence the area and power of our bespoke co-processor, this variation is relatively constrained for a fixed number of multipliers and can be largely ignored during constant selection.
Instead, to design the most energy-efficient system, we should focus on selecting constants that minimize the required decompositions, thereby reducing system latency and improving overall energy efficiency.

\section{Latency-aware Co-processor Optimization}~\label{sec:optimization}
\setlength{\parskip}{0pt}  
Employing a bespoke co-processor can improve metrics such as area and power.
However, the overall system performance and energy efficiency will be primarily determined by the choice of constants for the bespoke co-processor and by how the required operations (e.g., products) are executed over the available bespoke multipliers.
This observation may also explain why, despite the wide adoption of bespoke design in small, application-specific flexible circuits~\cite{Ozer:Nature:2020, Ozer:DFLEPS2020,velu24,Mubarik:MICRO:2020:printedml,Armeniakos:DATE2022:axml,Armeniakos:TC2023:codesign,Kokkinis:TC:2024:enabling,Mrazerk:ICCAD2024,Afentaki:ICCAD23:hollistic,Afentaki:DATE2024:embedding}, our work is the first to design flexible bespoke accelerated \mbox{RISC-V} microprocessors, which are essential for implementing more complex systems in an area-efficient manner~\cite{raisiardali2025flexingriscvinstructionsubset,ozer2024bendable,Vergos:ISVLSI2025}.
\vspace{-0.1cm}
\subsection{Optimization Problem Definition}
Preserving security and privacy, all optimizations are performed post-training using only the model weights.
Within this setting, given a set of weighted sums (e.g., the neurons in an MLP), our goal is to identify a set of constants that minimize the total latency of folding all required weighted sums on a bespoke co-processor configured with these constants, while also accounting for the cost of products that are not directly represented within each sum.
Since, as discussed in Section~\ref{sec:introduction}, we target ML classification using MLPs, we hereafter adopt MLP-specific terminology, though our approach can be seamlessly extended to any ML classifier that relies on MAC computations.
The objective of energy-efficient co-processor design can be formulated as an optimization problem that selects the optimal set of constants and corresponding product decompositions to minimize the total number of cycles per inference (i.e., latency, since the co-processor operates at the same clock frequency as SERV regardless of the selected constants) while respecting hardware constraints.
We reiterate that, for each target MLP, a unique bespoke co-processor is designed.
Formally, the optimization problem is defined as:
\begin{equation}
\begin{gathered}
\mathcal{CSPD} : (N, W, C) \;\longmapsto\; (C',D,  L):\quad \min_{C',D}L,\\
\text{subject to}\, \textit{multipliers constraints}
\label{eq:optimization}
\end{gathered}
\end{equation} 
where $N$ is the computational graph of the input MLP, consisting of all neurons.
$W = \{\mathbf{w}_1, \mathbf{w}_2, \dots, \mathbf{w}_r\}$ is the set of weight vectors, with each $\mathbf{w}_n$ representing the weights associated with neuron $n \in N$.
$C = \{c_1, c_2, \dots, c_m\}$ denotes the set of candidate constants for the bespoke multipliers.
The solution $C' = \{(c, k_c) \mid c \in C\}$ specifies the selected constants $c$ along with their replication factors $k_c$, i.e., the number of bespoke multipliers instantiated for each constant.
Finally, $D$ denotes the decompositions and $L$ the total inference latency.

The problem is structured into two tightly coupled stages:
i) from the candidate set $C$, a set $C'$ is selected to instantiate the bespoke multipliers in hardware, and
ii) given this configuration $C'$, the operations in $N$ are scheduled onto the available multipliers while respecting dependencies and minimizing overall latency.
The bespoke multipliers can implement weights in $W$ either directly or through decomposition, meaning that the joint choice of constants and decomposition decisions ultimately determines the system latency $L$.
Thus, solving $\mathcal{CSPD}$ requires finding both an optimal set of constants, along with a valid scheduling strategy, to minimize $L$.

\begin{algorithm}[t!]
\footnotesize
\caption{Constraint Programming Model}
\label{alg:weight-scheduling}
\begin{algorithmic}[1]
\State \textbf{Inputs:}
\begin{itemize}
    \item $N$: Set of neurons
    \item $W$: Set of weight vectors for each neuron $n\in N$
    \begin{itemize}
        \item $\text{w}_{n,i}$: the $i$-th weight of neuron $n$
    \end{itemize}
    \item $C$: Set of candidate constants (e.g., $[-8, 7]$)
    \item $M_{\text{total}}$: Maximum number of bespoke multipliers
    \item $T = \{1, \dots, T_{\max}\}$: Set of available scheduling cycles. $T_{\max}$ can be used as the latency constraint or set to a large value to always enable a solution.

\end{itemize}

\State \textbf{Decision Variables:}
\begin{itemize}
    \item $k_c$: Number of multipliers of type $c$
    \item $d_{n,t,i,c}$: Number of multipliers of type $c$ used in cycle $t$ for $\text{w}_{n,i}$
    \item $l_n$: Latency of neuron $n$
\end{itemize}

\State \textbf{Latency Definition:} 
$l_n = \max \{\, t \in T \mid \exists i,c: d_{n,t,i,c} > 0 \,\}$

\State \textbf{Objective:} 
$\min \sum_{n \in N} l_n$

\State \textbf{Constraints:}
\begin{itemize}
    \item \textbf{Total multiplier budget:} 
    $\sum_{c \in C} k_c \leq M_{\text{total}}$
    \item \textbf{Cycle capacity:} 
    $\sum_{i} d_{n,t,i,c} \leq k_c, 
    \quad \forall n \in N, \forall t \in T, \forall c \in C$
    \item \textbf{Decomposition:} 
    $\sum_{t \in T} \sum_{c \in C} d_{n,t,i,c} \cdot c = \text{w}_{n,i}, 
    \quad \forall n \in N, \forall i$
\end{itemize}
\end{algorithmic}
\end{algorithm}

\vspace{-0.1cm}
\subsection{Constraint Programming-Based Solution}

To solve~\eqref{eq:optimization}, we employ the \textit{CP-SAT} solver from Google’s OR-Tools~\cite{cpsatlp}, an open-source constraint programming engine consistently ranking among the top performers in the \textit{MiniZinc Challenge}~\cite{Stuckey_Feydy_Schutt_Tack_Fischer_2014,minizinc_challenge}.
The solver combines Boolean satisfiability (SAT) solving with constraint-programming-style propagation and learning, enabling efficient pruning of infeasible solutions through conflict analysis and constraint propagation.

We model this problem as a constraint instance by defining:
(i) decision variables that capture the design space,
(ii) constraints that enforce hardware and dependency requirements, and
(iii) an objective function that drives the minimization of total inference latency.
The formal formulation for neuron weight scheduling is presented in Algorithm~\ref{alg:weight-scheduling}.
Here, the total multiplier budget $M_{\text{total}}$ enforces hardware limits on the number of bespoke multipliers, while the cycle capacity constraint prevents resource overuse in any given cycle.
The product (weight) decomposition constraint ensures computational correctness. Decomposition may map a weight directly to an available multiplier with a matching constant or partition it when no exact match exists or when the matching multiplier is already in use, with partitioning possibly unfolding over multiple cycles.
Finally, the objective function minimizes the sum of neuron latencies, steering the solver toward minimal total inference time.
Feasibility is ensured as, for sufficiently large execution time (large $T_{\max}$), one $+1$ and one $-1$ bespoke multiplier suffice to realize any weight via decomposition.

CP-SAT with Algorithm~\ref{alg:weight-scheduling} explicitly solves the decomposition problem and the variables $d_{n,t,i,c}$ (i.e., extracted decompositions for each weight of every neuron) are used to automatically generate the co-processor calls in software.
However, the set $C'$ is also implicitly determined.
All constants $c \in C$ with $k_c \neq 0$ form the final set $C'$ and define the constant values hardwired into the bespoke multipliers.

Given the identified set $C'$, the Verilog description of our co-processor is automatically generated from our parameterized templates and seamlessly integrated with the SERV core.
As concluded in Section~\ref{sec:bespoke}, the specific constant values have less impact on overall system efficiency.
Therefore, the candidate set $C$ is populated according to the weight quantization scheme.
For instance, for $4$-bit quantized MLP weights, we set $C = [-8, 7]$.
In a more restrictive case, $C$ can be set to hardware-friendly constants only (see Fig.~\ref{fig:bespoke_mul}).
Finally, since \textit{CP-SAT} solves only single-objective optimization problems, we may explore different trade-offs by running the solver for multiple values of $M_{\text{total}}$.
With $4$-bit inputs we run the solver for $M_{\text{total}} = 16$.
Exploring smaller $M_{\text{total}}$ values may optimize energy consumption alongside area and power (see Fig.~\ref{fig:mc}).

\begin{figure}[!t]   \includegraphics[width=\linewidth]{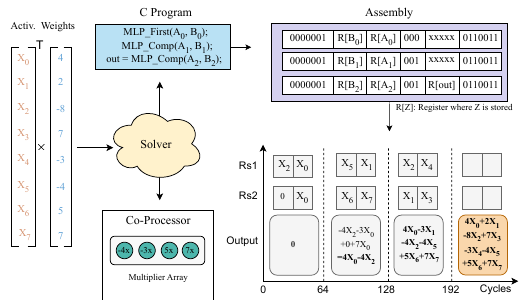}
    \caption{Cycle-level example of neuron output accumulation.}
    \label{fig:solver-exam}
    \vspace{-3ex}
\end{figure}

Fig.~\ref{fig:solver-exam} presents an illustrative decomposition example.
For simplicity, inputs are assumed to be $16$-bit (i.e., two inputs per register), and the co-processor features four bespoke multipliers.
The solver produces two outputs: (i) the configuration of the by-constant multipliers, which is directly reflected in the generated co-processor, and (ii) a schedule list of decompositions, indicating which products must be computed at each co-processor call (see Fig.~\ref{fig:solver-exam}) by assigning the respective activations to the corresponding register positions.

\section{Results and Analysis}
\setlength{\parskip}{0pt}   
\subsection{Experimental Setup}
\label{subsec:model_selection}

We evaluate our framework on diverse popular healthcare benchmarks (Table~\ref{tab:dataset_accuracy}) spanning stress monitoring, cardiology, dermatology, activity recognition, and emotion-aware driving. Specifically, we use WESAD~\cite{wesad}, DriveDB~\cite{drive}, StressInNurses~\cite{stressinnurses}, and SPD~\cite{spd} for stress detection; Arrhythmia~\cite{arrhythmia} and ECG5000~\cite{ecg5000} for cardiac analysis; Dermatology~\cite{dermatology} for skin disease classification; HAR~\cite{HAR} for activity recognition; and AffectiveRoad~\cite{affectiveroad} for driving-related emotion recognition.
Each dataset is split into training and testing sets with an $70$/$30$ ratio, normalized to $[-1,1]$, and used to train MLPs in TensorFlow Lite.
Considering the strict constraints of flexible electronics, we limit MLPs to a maximum of two hidden layers with up to ten neurons per layer and apply quantization and pruning during training to produce compact models that meet the requirements of flexible applications.
For each dataset, we perform a neural architecture search to identify the smallest network that achieves top accuracy, using $4$-bit weights and either $4$- or $5$-bit activations for quantization.
Table~\ref{tab:dataset_accuracy} reports the extracted MLP topologies and their achieved accuracies.
As shown, the optimized FlexIC-oriented MLPs incur at most $0.75$\% accuracy loss compared to the unpruned floating-point model.


\begin{table}[t]
\centering
\setlength{\tabcolsep}{2pt}
\caption{Accuracy Evaluation}
\label{tab:dataset_accuracy}
\begin{tabular}{lcccc}
\toprule
\textbf{Dataset} & \textbf{Topology} & \textbf{\#MAC} & \textbf{Acc. FP32$^1$(\%)} & \textbf{Acc. QP$^2$(\%)} \\
\midrule
AffectiveRoad   & $63$-$9$-$3$     & $594$  & $97.68$ & $97.17$ \\
Arrhythmia       & $279$-$9$-$11$   & $2610$ & $70.00$ & $70.00$ \\
Dermatology      & $34$-$9$-$6$     & $360$  & $95.06$ & $94.44$ \\
DriveDB          & $61$-$9$-$3$     & $576$  & $93.06$ & $92.51$ \\
ECG5000          & $140$-$3$-$5$    & $435$  & $92.84$ & $92.10$ \\
HAR              & $561$-$7$-$6$    & $3969$ & $93.72$ & $93.18$ \\
SPD$^3$          & $75$-$9$-$3$     & $702$  & $90.75$ & $90.05$ \\
StressInNurses   & $72$-$9$-$3$     & $675$  & $74.11$ & $73.36$ \\
WESAD            & $96$-$9$-$3$     & $891$  & $82.56$ & $82.46$ \\
\bottomrule
\end{tabular}
\par\vspace{1ex}
\footnotesize 
$^1$Accuracy Floating-point model. $^2$Accuracy optimized model. \\
$^3$SPD required 5-bit activations, while all other MLPs use 4-bit ones.
\vspace{-3ex}
\end{table}

We perform full-chip design (from RTL to GDSII) of our ML-accelerated microprocessors.
Synopsys Design Compiler S-2021.06 is used for synthesis and mapping to the FlexIC Gen3 $3$V standard-cell library~\cite{flexic_gen3}.
Synopsys VCS T-2022.06 is used for timing simulations, while PrimeTime T-2022.03 for timing and power analysis.
Physical implementation is carried out using Cadence Innovus 21.35, where we design the chip floorplan and power delivery network, followed by placement, clock tree synthesis, routing, and post-routing optimizations.
Finally, timing and power sign-off tools are used to accurately report the chip's delay and power after parasitic extraction.
All syntheses and evaluations are conducted at $150$kHz, as at this frequency SERV provided a good area and power trade-off. 

In our analysis, we design and evaluate one flexible chip per MLP.
The bespoke co-processor is generated using our automated optimization flow (Section~\ref{sec:optimization}), while the SERV core is based on the FlexIC-modified version of~\cite{ozer:nature2024:bendableRiscV}, which implements the RV32E ISA, reduces the register file to 16 registers, and employs a custom serial peripheral interface (SPI) to access external memory, overcoming technology limitations but incurring a 46/47-cycle latency per read/write.

For comparisons, we perform a full-chip design of Flex-RV~\cite{ozer:nature2024:bendableRiscV}, which integrates a generic, conventional ML co-processor featuring two $8$x$4$ and two $4$x$4$ multipliers, along with dedicated post-processing hardware for bias addition, and ReLU activation.
Additionally, we design a Semi-bespoke system that uses the same SERV core but couples it with a co-processor customized only to the precision required by each trained model.
Specifically, for SPD, the Semi-bespoke co-processor integrates seven $5$x$4$ conventional multipliers, while for the other MLPs it uses eight $4$x$4$ conventional multipliers.

Finally, all systems are also prototyped on a Digilent Arty A7-50T FPGA board.
Bitstreams are generated using Vivado 2020.1 with default synthesis and place-and-route constraints.
Each FPGA prototype is then validated by running inference on the test dataset to verify correct operation.

\begin{table}[t!]
\centering
\footnotesize
\setlength{\tabcolsep}{2pt}
\caption{Post-layout evaluation. Target frequency $150$kHz.}
\label{tab:postlayout}
\hspace{-7mm}
\begin{tabular}{c@{}c l|ccc|cccc}
\cmidrule(lr){2-10}
& & \multirow{4}{*}{\textbf{Design}}
& \multicolumn{3}{c|}{\textbf{Total}}
& \multicolumn{4}{c}{\textbf{Co-processor}} \\
\cmidrule(lr){4-6} \cmidrule(lr){7-10}
& & & \thead{\textbf{Area}\\(mm$^2$)}& \thead{\textbf{Power}\\(mW)} & \textbf{\#Gates}
& \thead{\textbf{Area}\\(mm$^2$)}& \thead{\textbf{Power}\\(mW)} & \thead{\textbf{Delay}\\(\si{\micro\second})} & \textbf{\#Gates} \\
\cmidrule(lr){2-10}
\multirow{9}{2mm}{\rotatebox{90}{Ours}} & \ldelim\{{9}{2mm} & AffectiveRoad    & $2.433$ & $1.511$ & $4235$ & $0.181$ & $0.144$ & $3.9$ & $585$ \\
& & Arrhythmia         & $2.410$ & $1.493$ & $4140$ & $0.158$ & $0.126$ & $3.6$ & $490$ \\
& & Dermatology        & $2.417$ & $1.504$ & $4183$ & $0.165$ & $0.137$ & $3.6$ & $533$ \\
& & DriveDB            & $2.442$ & $1.527$ & $4291$ & $0.190$ & $0.160$ & $3.8$ & $641$ \\
& & ECG5000            & $2.432$ & $1.519$ & $4244$ & $0.180$ & $0.152$ & $3.9$ & $594$ \\
& & HAR                & $2.425$ & $1.506$ & $4208$ & $0.173$ & $0.139$ & $3.6$ & $558$ \\
& & SPD$^\dagger$                & $2.413$ & $1.498$ & $4172$ & $0.161$ & $0.131$ & $3.9$ & $522$ \\
& & StressInNurses     & $2.445$ & $1.523$ & $4290$ & $0.193$ & $0.156$ & $3.8$ & $640$ \\
& & WESAD              & $2.443$ & $1.521$ & $4281$ & $0.191$ & $0.154$ & $3.8$ & $631$ \\
\cmidrule(lr){2-10}
\multirow{4}{2mm}{\rotatebox{90}{SoA}} & \ldelim\{{4}{2mm}  & SERV$^\S$               & $2.252$ & $1.367$ & $3650$ & -- & -- & -- & -- \\
& & Flex-RV~\cite{ozer:nature2024:bendableRiscV} & $2.542$ & $1.627$ & $4531$ & $0.290$ & $0.260$ & $4.8$ & $881$ \\
& & SB$^*$ w/ $4$x$4$ x$8$     & $2.492$ & $1.587$ & $4635$ & $0.240$ & $0.220$ & $3.7$ & $985$ \\
& & SB$^*$ w/ $5$x$4$ x$7$    & $2.476$ & $1.619$ & $4628$ & $0.224$ & $0.252$ & $4.0$ & $978$ \\
\cmidrule(lr){2-10}
\end{tabular}
\raggedleft\par\noindent\footnotesize $^\dagger$SPD uses $12$ $5$-bit bespoke multipliers while all others use $16$ $4$-bit ones.
$^\S$SERV delay is $5.3$\si{\micro\second}. $^*$Semi-bespoke. 
\vspace{-1ex}
\end{table}

\begin{figure}[t!] \centering \begin{overpic}[trim=2.8cm 0.5cm 2.8cm 0.5cm,clip=true,width=0.25\columnwidth]{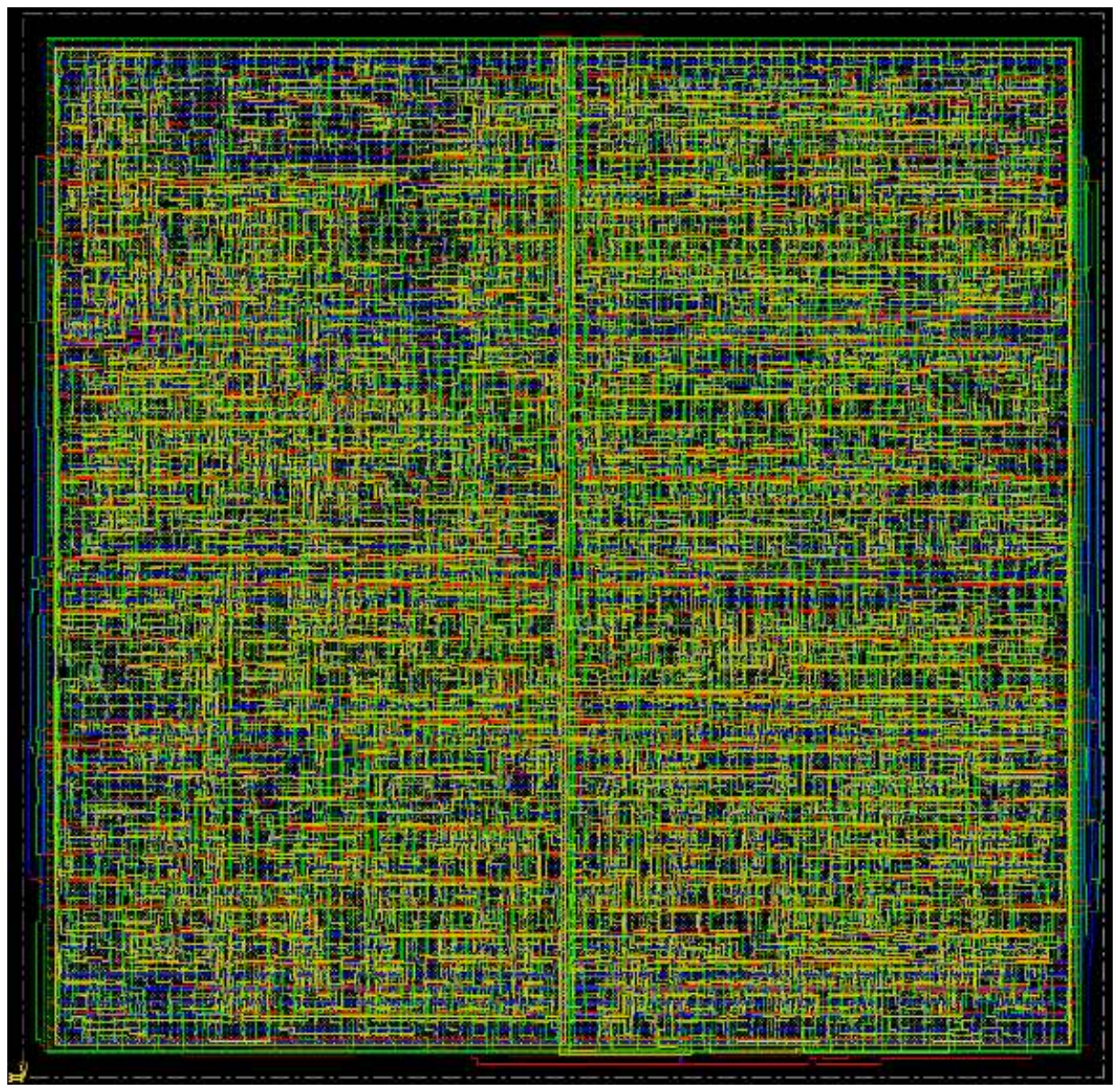} \subfiglabel{a} \end{overpic}\hspace{-8pt} \begin{overpic}[trim=2.8cm 0.5cm 2.8cm 0.5cm,clip=true,width=0.25\columnwidth]{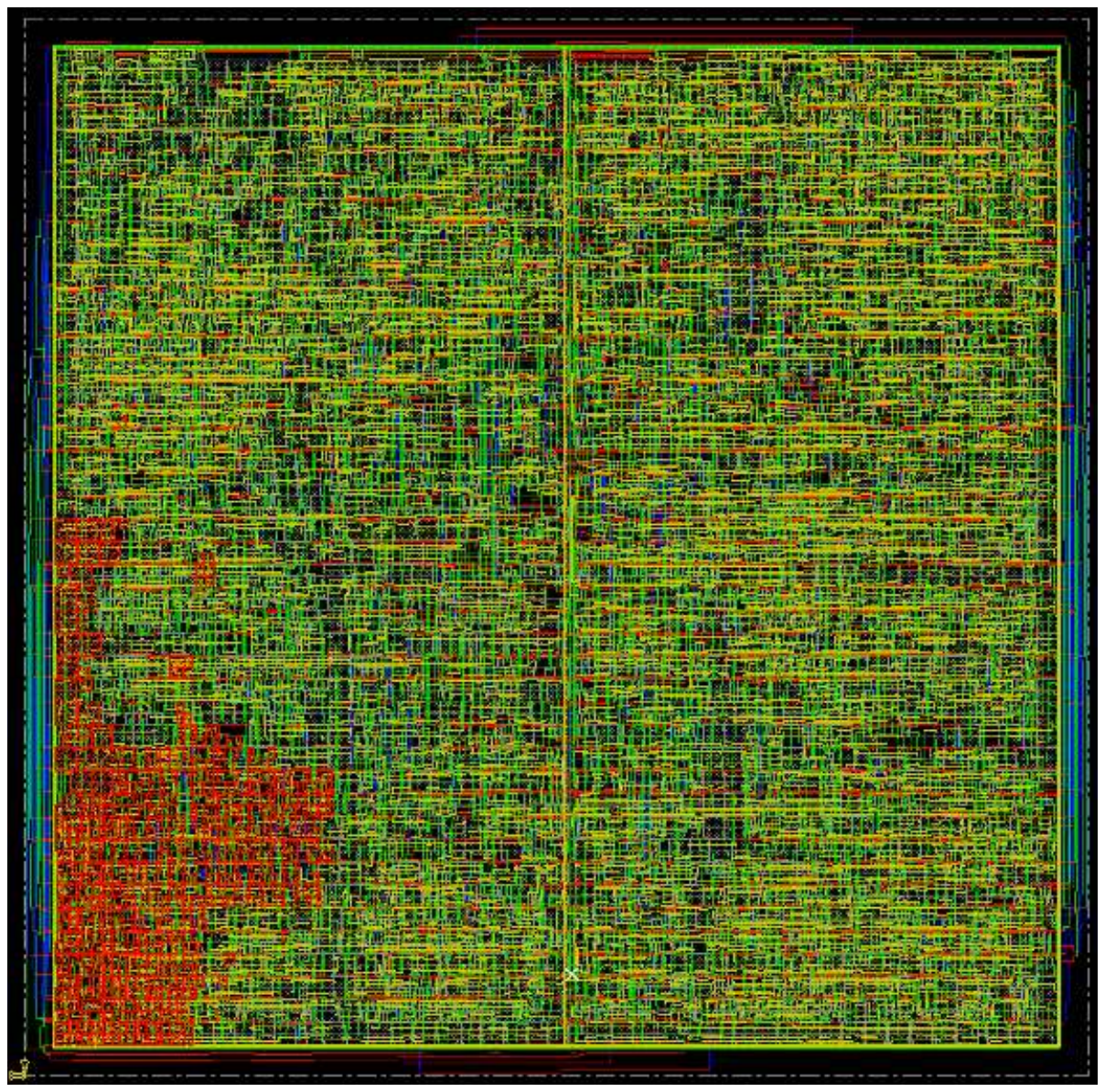} \subfiglabel{d} \end{overpic}\hspace{-8pt} \begin{overpic}[trim=2.8cm 0.5cm 2.8cm 0.5cm,clip=true,width=0.25\columnwidth]{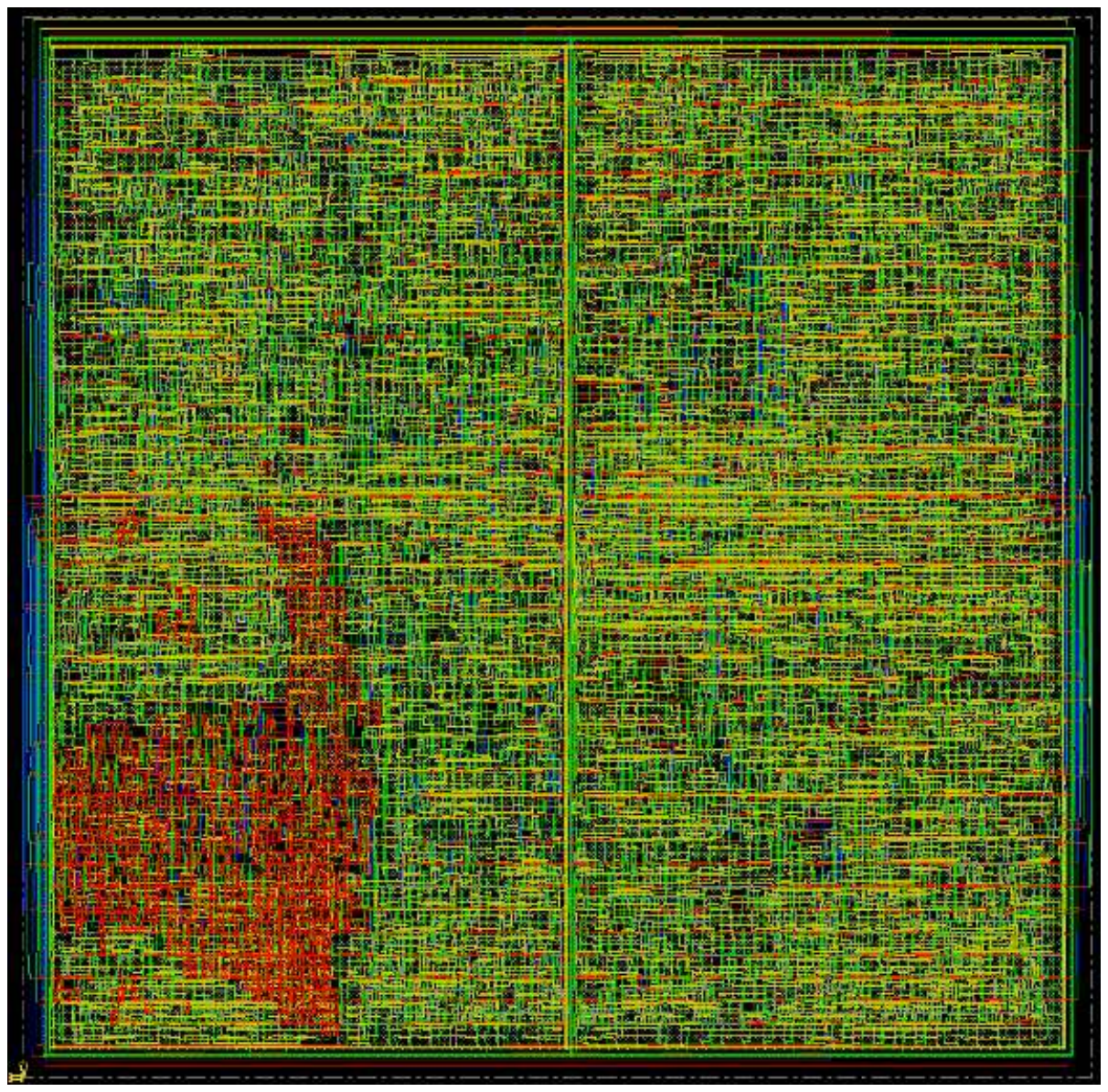} \subfiglabel{g} \end{overpic}\hspace{-8pt} \begin{overpic}[trim=2.8cm 0.5cm 2.8cm 0.5cm,clip=true,width=0.25\columnwidth]{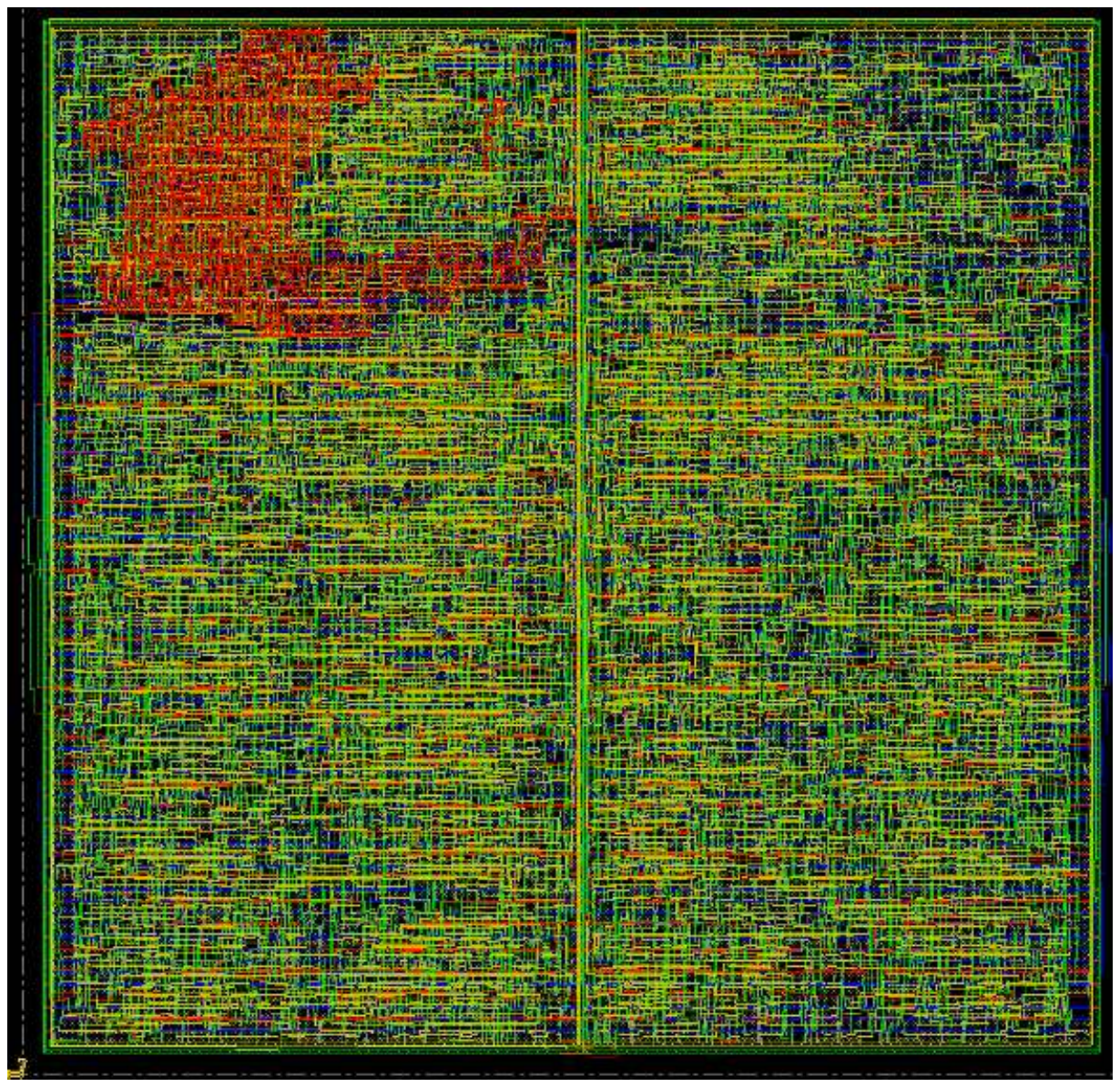} \subfiglabel{j} \end{overpic}\vspace{2mm} \begin{overpic}[trim=2.8cm 0.5cm 2.8cm 0.5cm,clip=true,width=0.25\columnwidth]{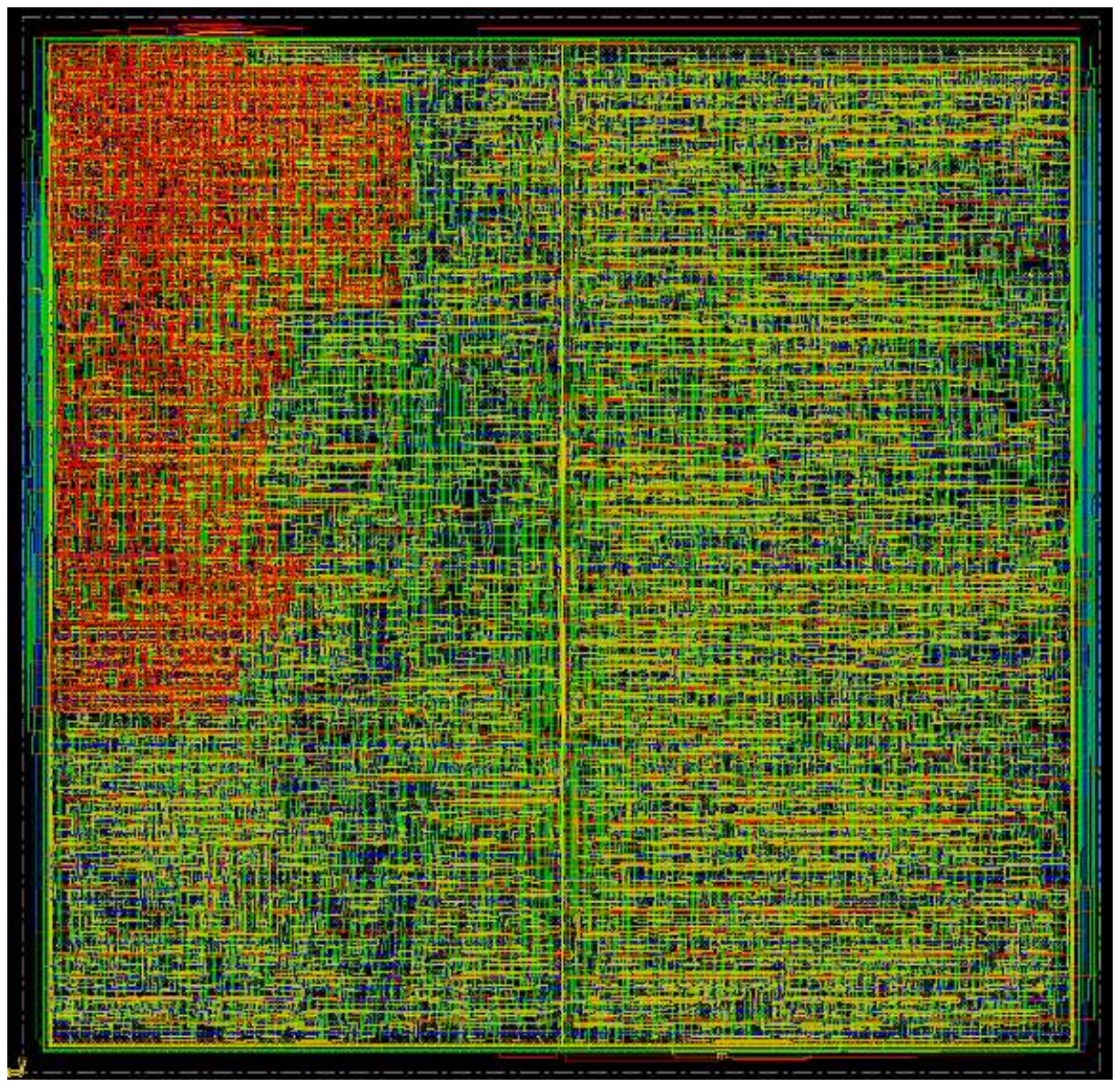} \subfiglabel{b} \end{overpic}\hspace{-8pt} \begin{overpic}[trim=2.8cm 0.5cm 2.8cm 0.5cm,clip=true,width=0.25\columnwidth]{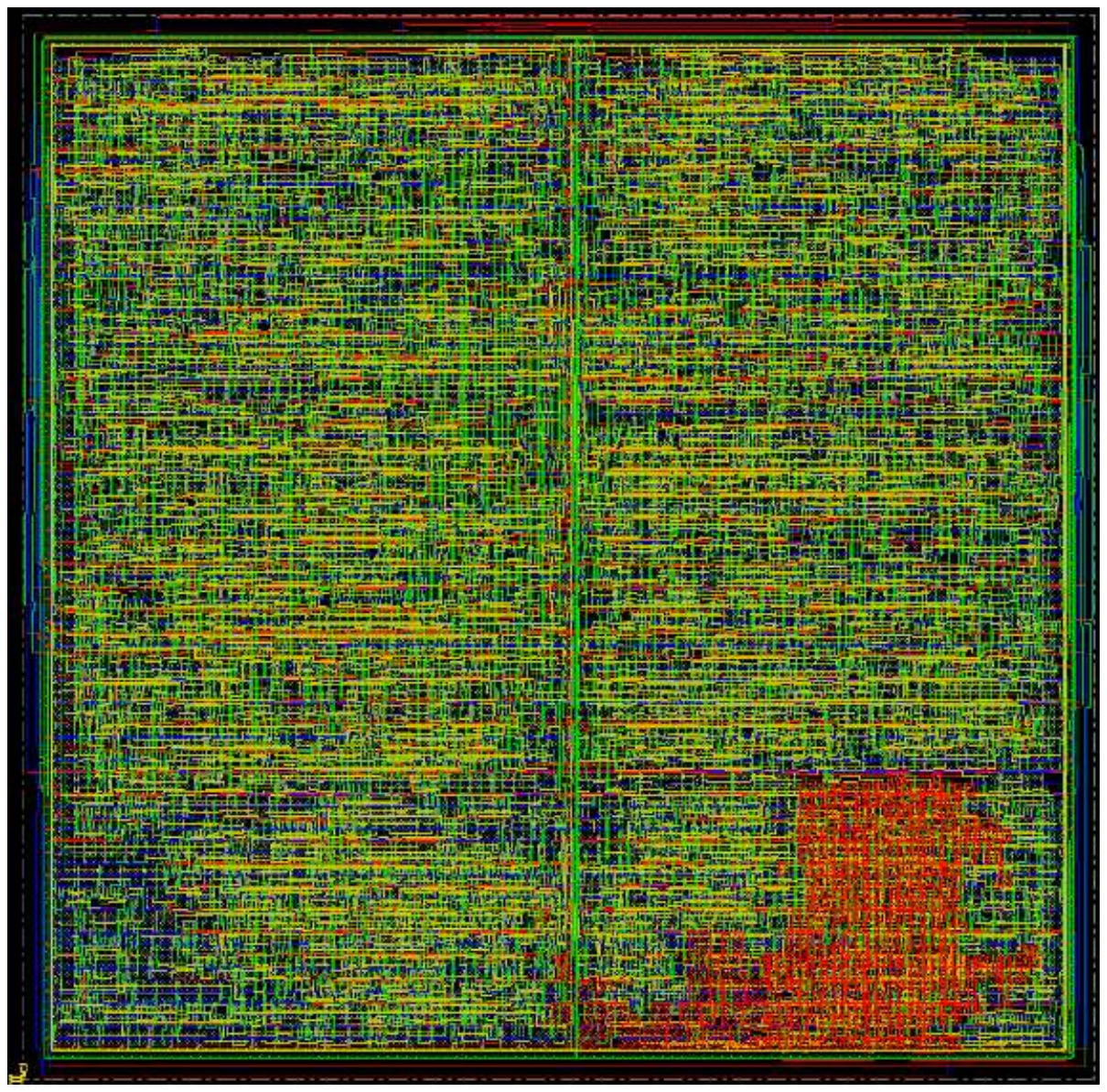} \subfiglabel{e} \end{overpic}\hspace{-8pt} \begin{overpic}[trim=2.8cm 0.5cm 2.8cm 0.5cm,clip=true,width=0.25\columnwidth]{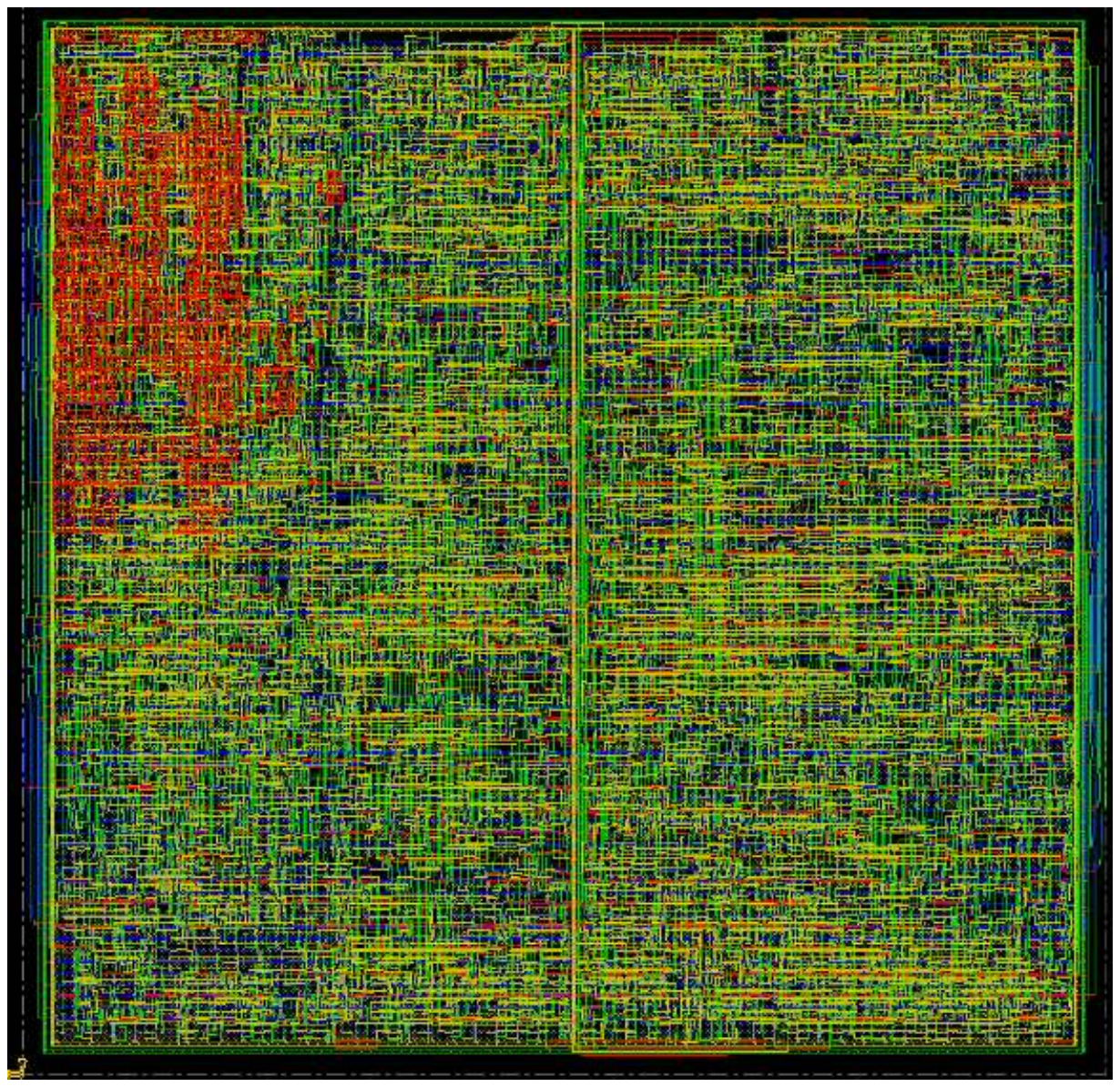} \subfiglabel{h} \end{overpic}\hspace{-8pt} \begin{overpic}[trim=2.8cm 0.5cm 2.8cm 0.5cm,clip=true,width=0.25\columnwidth]{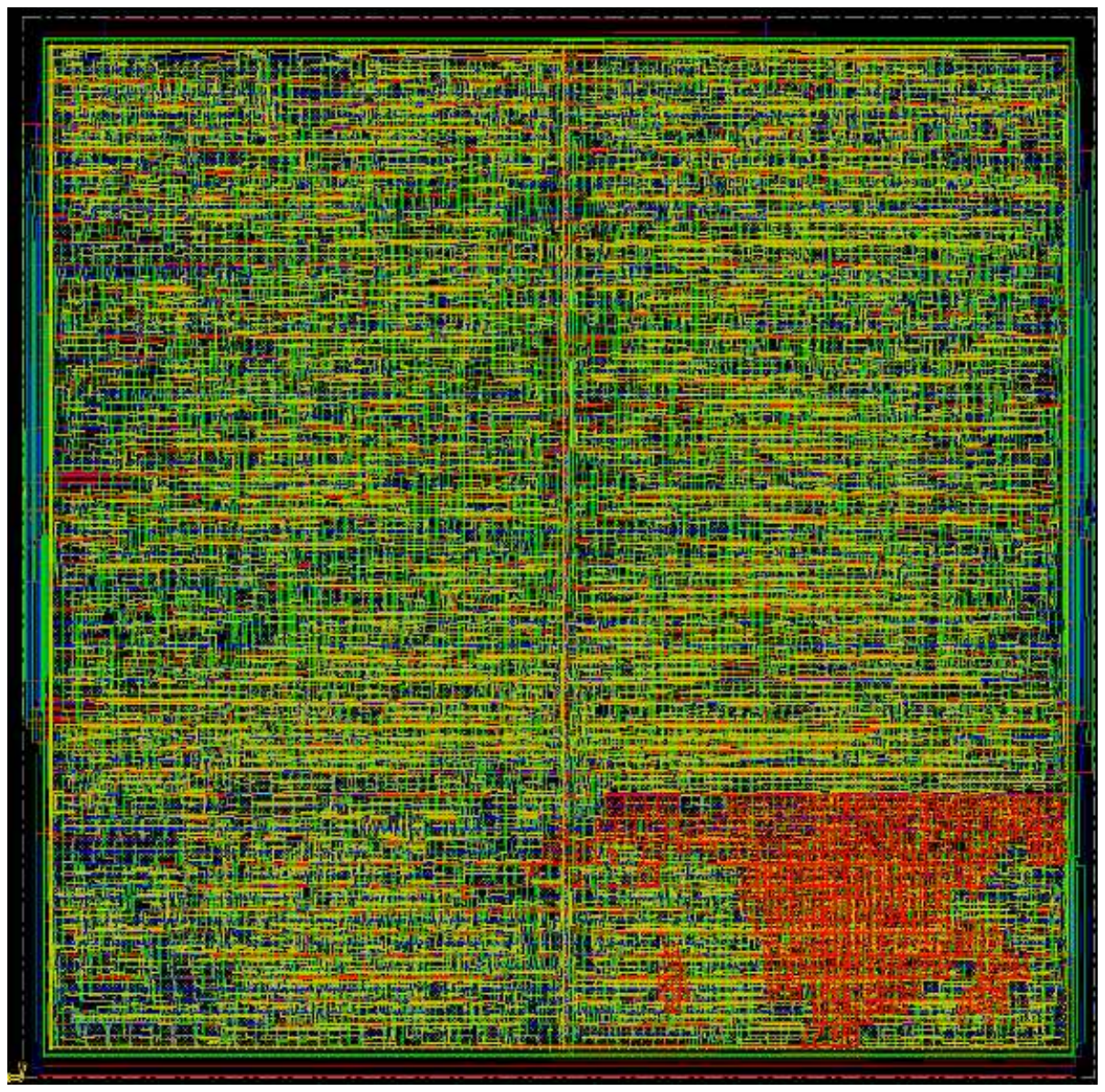} \subfiglabel{k} \end{overpic}\vspace{2mm} \begin{overpic}[trim=2.8cm 0.5cm 2.8cm 0.5cm,clip=true,width=0.25\columnwidth]{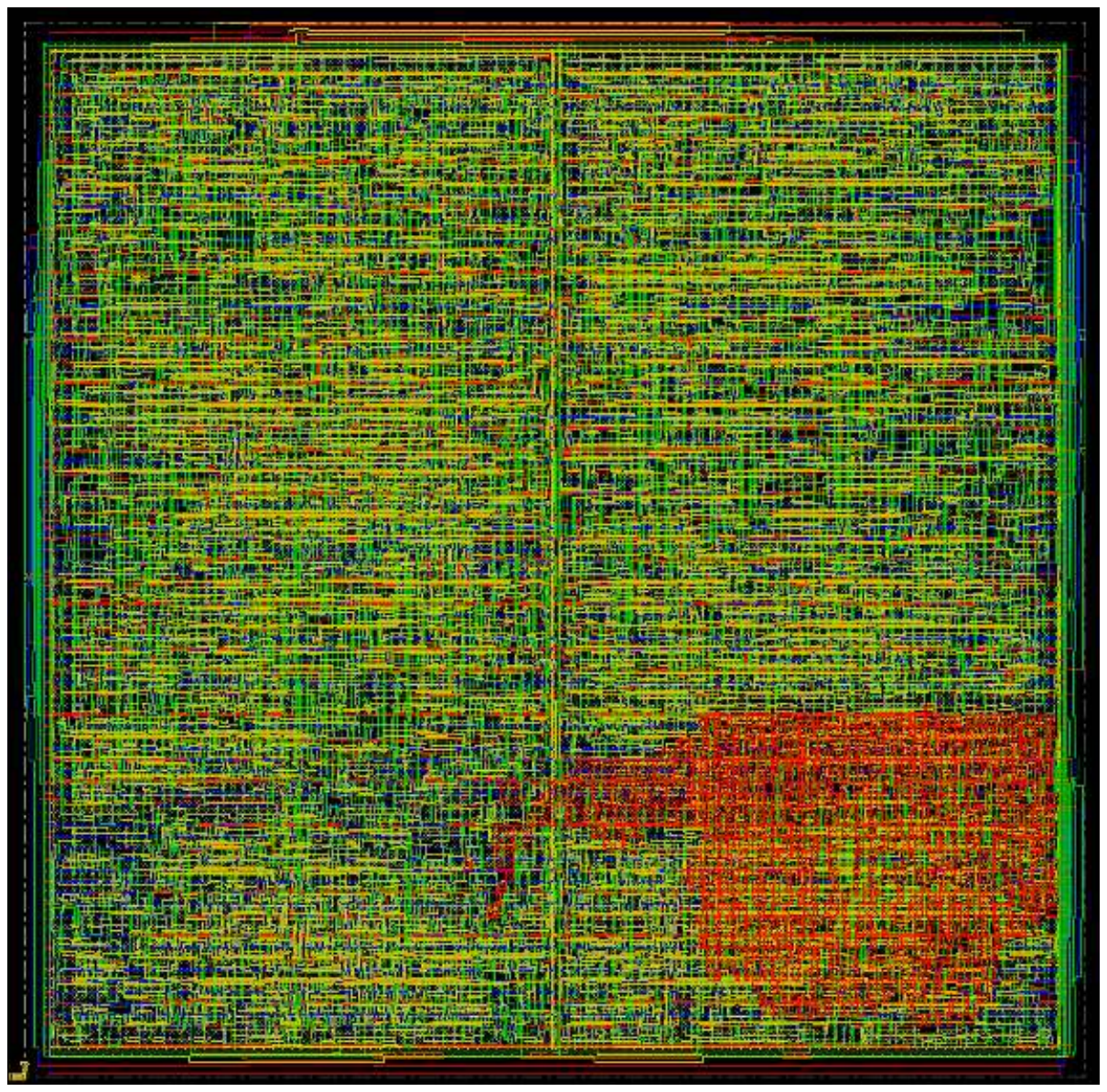} \subfiglabel{c} \end{overpic}\hspace{-8pt} \begin{overpic}[trim=2.8cm 0.5cm 2.8cm 0.5cm,clip=true,width=0.25\columnwidth]{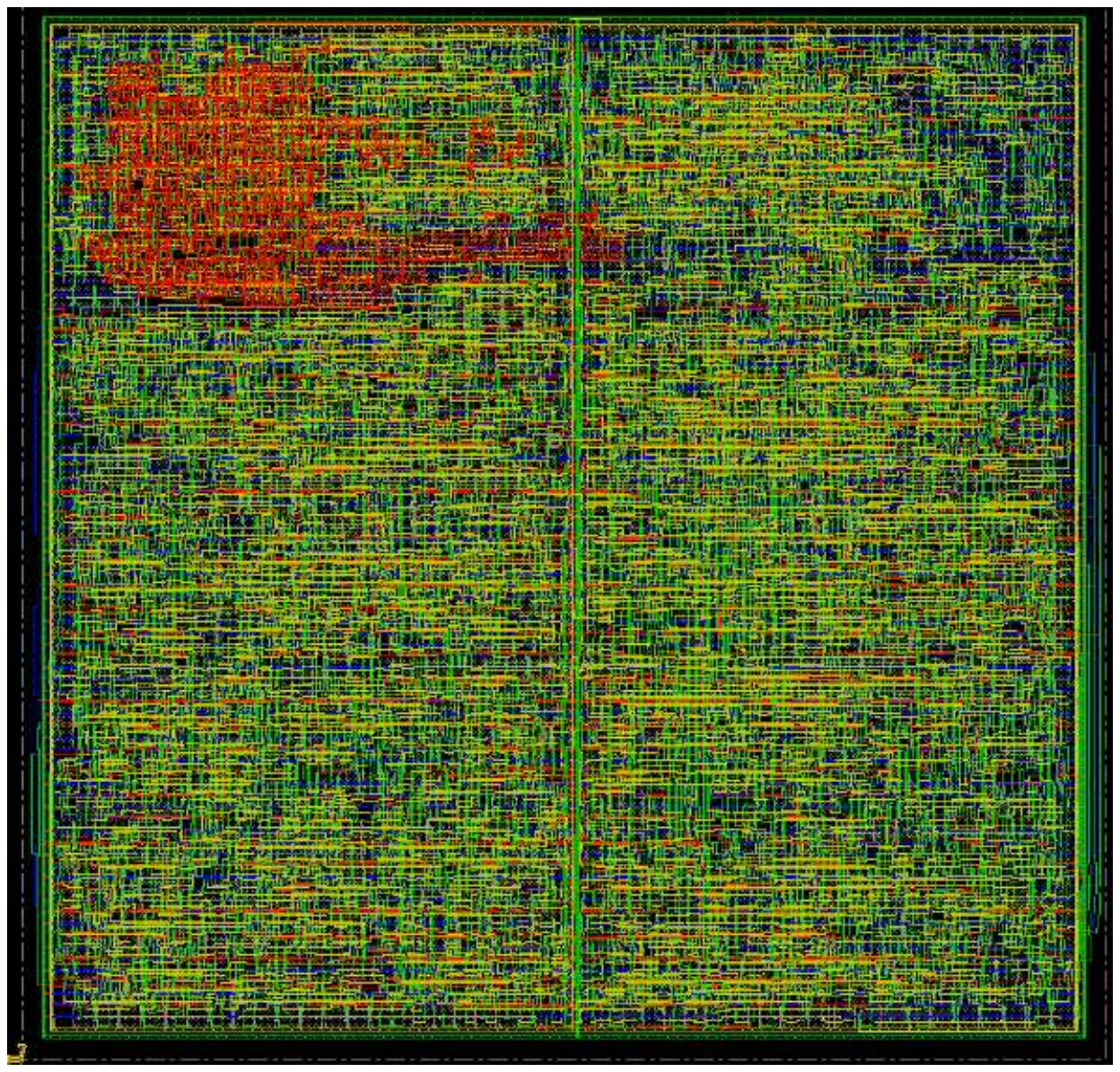} \subfiglabel{f} \end{overpic}\hspace{-8pt} \begin{overpic}[trim=2.8cm 0.5cm 2.8cm 0.5cm,clip=true,width=0.25\columnwidth]{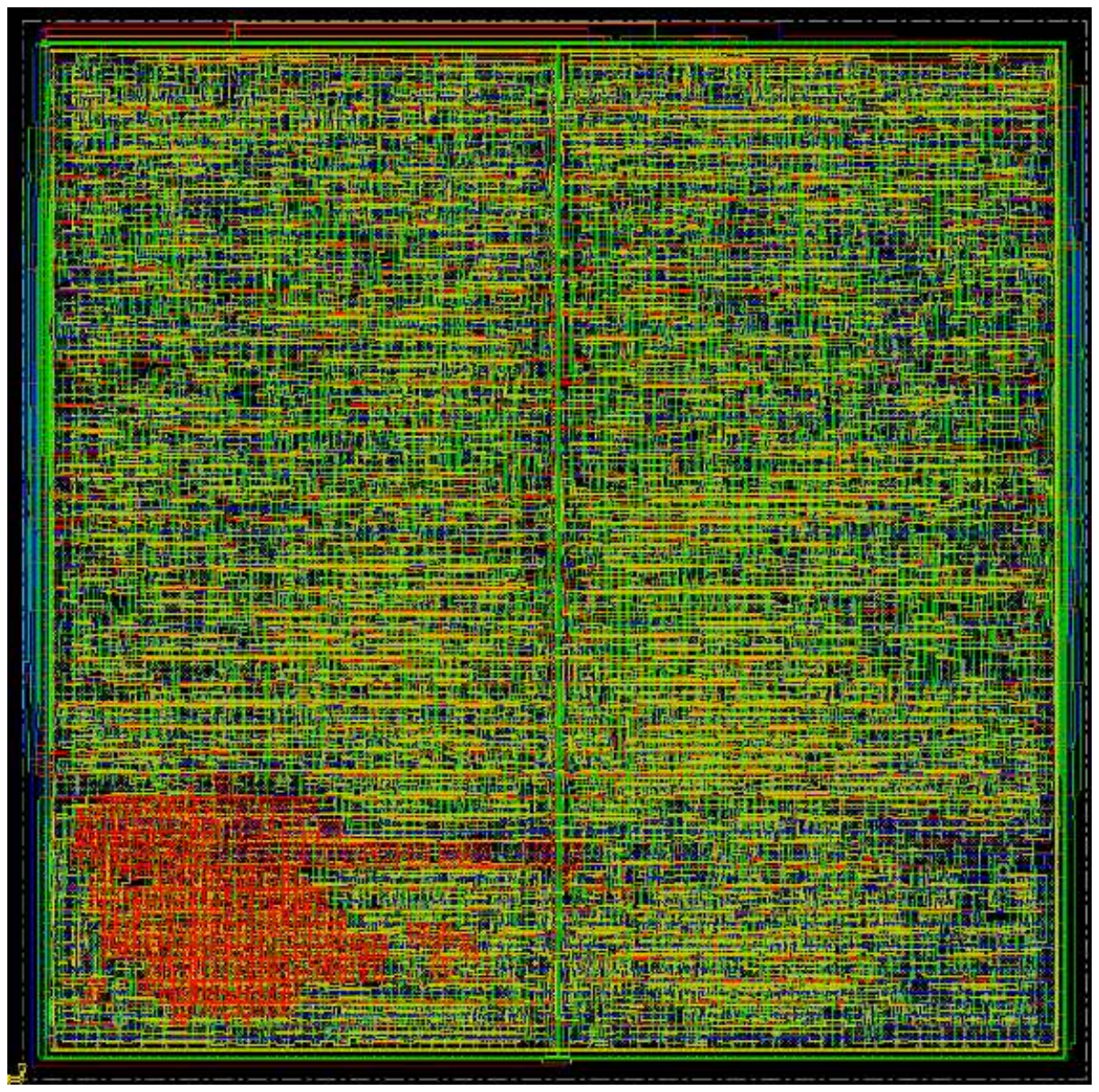} \subfiglabel{i} \end{overpic}\hspace{-8pt} \begin{overpic}[trim=2.8cm 0.5cm 2.8cm 0.5cm,clip=true,width=0.25\columnwidth]{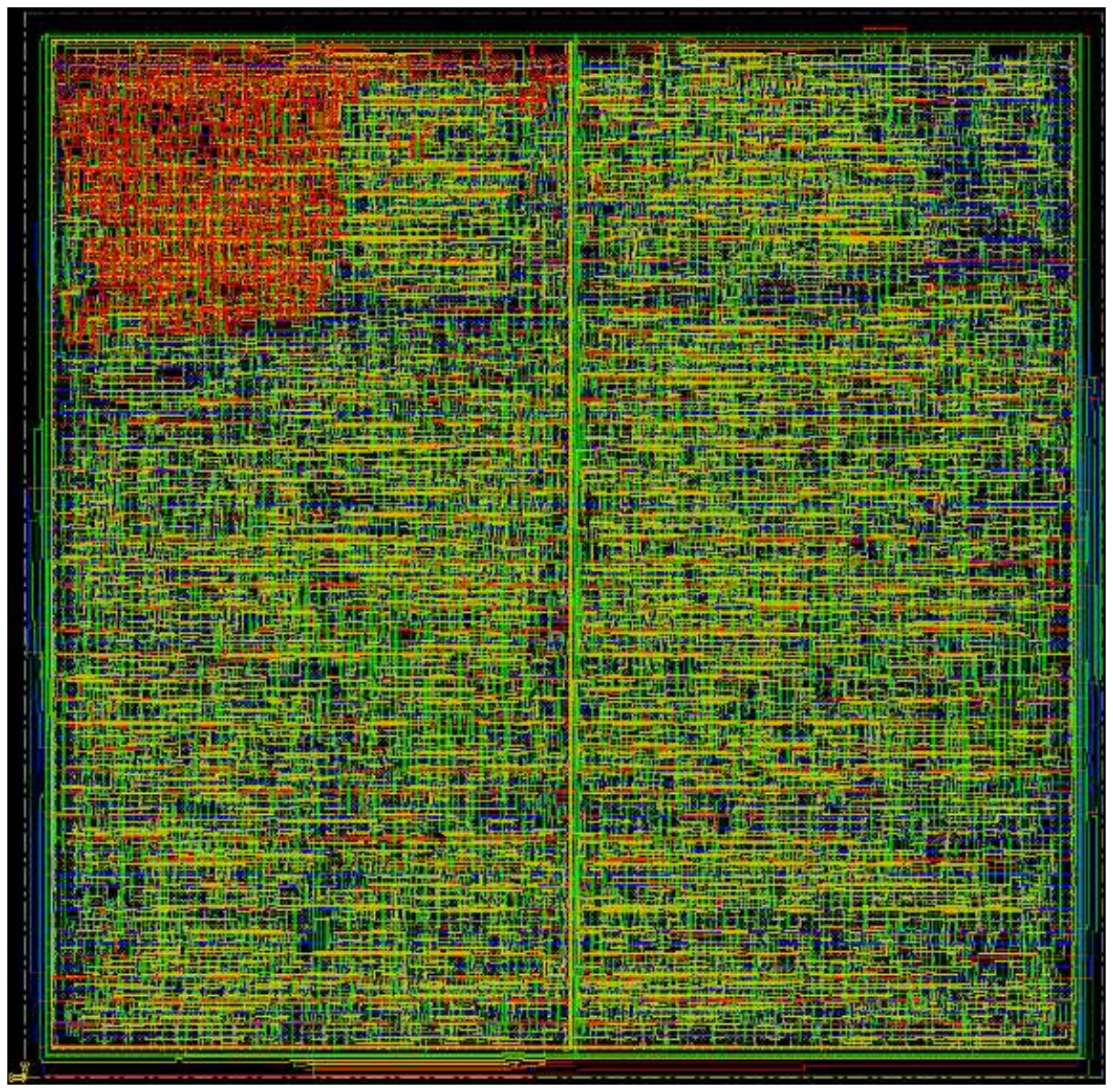} \subfiglabel{l} \end{overpic}
\noindent
\raisebox{3mm}{\(\underbrace{\hspace{0.2\columnwidth}}_{\text{SoA}}\)}%
\hspace{0.04\columnwidth}%
\raisebox{3mm}{\(\underbrace{\hspace{0.7\columnwidth}}_{\text{Ours}}\)}%
\caption{Layout of (a) SERV (non-accelerated), (b) Flex-RV~\cite{ozer:nature2024:bendableRiscV}, (c) Semi-bespoke w/ eight $4$x$4$ multipliers, and ours: (d) AffectiveRoad, (e) Arrhythmia, (f) Dermatology, (g) DriveDB, (h) ECG5000, (i) HAR, (j) SPD, (k) StressInNurses, and (l) WESAD. Red regions highlight the co-processor. } \label{fig:Layouts} \vspace{-3ex} \end{figure}

\vspace{-0.1cm}
\subsection{Evaluation}

Table~\ref{tab:postlayout} reports the hardware evaluation of the designed chips.
Compared to the vanilla, non-accelerated SERV, our co-processors introduce only a  $8$\% area and $10$\% power overhead, on average.
Still, all our chips maintain a compact footprint, averaging $2.42$mm$^2$, making them suitable for most flexible wearables, while their average power consumption of $1.5$mW allows them to be powered by existing flexible batteries or even flexible energy harvesters~\cite{Afentaki:DATE2024:embedding,Kokkinis:TC:2024:enabling}.
On average, our chips contain $4200$ NAND2-equivalent gates, ensuring realistic implementations and their co-processor critical path delay is at most $3.9$\si{\micro\second}, easily satisfying the $150$kHz target frequency. For reference, SERV’s critical path delay is $5.3$\si{\micro\second}.

Compared to Flex-RV~\cite{ozer:nature2024:bendableRiscV} and Semi-bespoke, our chips are slightly smaller and more power-efficient.
This is explained by the fact that compared to Flex-RV~\cite{ozer:nature2024:bendableRiscV}, our co-processor is on average $39$\% smaller and consumes $45$\% less power.
Similarly, against Semi-bespoke, our co-processor is on average $26$\% smaller with $36$\% lower power consumption.
Finally, Fig.~\ref{fig:Layouts} illustrates the layouts for SERV, Flex-RV~\cite{ozer:nature2024:bendableRiscV}, Semi-bespoke with eight 4x4 conventional multipliers, and our chips.

\begin{table}[t] 
\centering
\footnotesize
\setlength{\tabcolsep}{3pt}
\caption{Inference latency and enegy efficiency evaluation} 
\label{tab:dataset_performance} 
\begin{tabular}{l| c c c c | c c c c}
\toprule
\multirow{3}{*}{\textbf{Dataset}}
& \multicolumn{4}{c|}{\textbf{Inference Latency (s)}}
& \multicolumn{4}{c}{\textbf{Energy/Inference (mJ)}} \\
\cmidrule(lr){2-5} \cmidrule(lr){6-9}
& \textbf{Ours} & \textbf{~\cite{ozer:nature2024:bendableRiscV}} & \textbf{SERV} & \textbf{SB}$^*$
& \textbf{Ours} & \textbf{~\cite{ozer:nature2024:bendableRiscV}} & \textbf{SERV} & \textbf{SB}$^*$ \\
\midrule
AffectiveRoad   & $0.639$ & $0.917$ & $647$   & $0.810$ & $0.51$ & $0.74$ & $786$   & $0.66$\\
Arrhythmia       & $0.999$ & $2.625$ & $2402$  & $1.558$ & $0.66$ & $1.72$ & $2889$  & $1.16$\\
Dermatology      & $0.504$ & $0.788$ & $404$   & $0.623$ & $0.45$ & $0.64$ & $489$   & $0.60$\\
DriveDB          & $0.581$ & $1.042$ & $624$   & $0.813$ & $0.47$ & $0.82$ & $757$   & $0.80$\\
ECG5000          & $0.212$ & $0.499$ & $437$   & $0.434$ & $0.18$ & $0.33$ & $526$   & $0.36$\\
HAR              & $0.638$ & $3.586$ & $4492$  & $2.082$ & $0.45$ & $2.30$ & $5446$  & $1.37$\\
SPD              & $0.756$ & $1.675$ & $654$   & $0.976$ & $0.63$ & $1.15$ & $790$   & $0.82$\\
StressInNurses   & $0.598$ & $0.965$ & $620$   & $0.726$ & $0.50$ & $0.78$ & $747$   & $0.66$\\
WESAD         & $0.703$ & $1.174$ & $817$   & $0.924$ & $0.50$ & $0.89$ & $985$   & $0.73$\\

\bottomrule
\end{tabular}
\raggedleft\par\vspace{1ex}\noindent\footnotesize $^*$Semi-bespoke
\vspace{-4ex}
\end{table}

Next, we evaluate ML inference on each flexible chip.
We run inference on the test datasets of the optimized models in Table~\ref{tab:dataset_accuracy} on our chips, as well as SERV, Flex-RV~\cite{ozer:nature2024:bendableRiscV}, and Semi-bespoke.
Table~\ref{tab:dataset_performance} reports the inference latency and the energy consumed per inference.
Latency is measured as the average cycles per sample multiplied by the clock period (i.e., $6.6$\si{\micro\second}), where cycle measurement includes memory accesses.
As shown in Table~\ref{tab:dataset_performance}, all our flexible chips achieve near-realtime operation, with latencies under $1$s.
In contrast, \mbox{Flex-RV} and Semi-bespoke violate this $1$s constraint in $55$\% and $22$\% of the cases, respectively.
More specifically, our chips deliver a $1972$x speedup over the non-accelerated SERV, $2.35$x over Flex-RV~\cite{ozer:nature2024:bendableRiscV}, and $1.6$x over Semi-bespoke.

Our chips consume, on average, as low as $0.48$mJ per inference, featuring $53.5$\% lower energy compared to the state-of-the-art Flex-RV~\cite{ozer:nature2024:bendableRiscV} and $39.3$\% versus the Semi-bespoke.
Energy gains are lower than latency gains, since, despite our co-processors being lower power and our chips achieving significantly lower latency, a substantial portion of execution time is spent on SERV due to its serial execution and memory accesses.
Still, our energy gains are substantial, making our circuit a highly energy-efficient and a promising solution for realistic ML-powered chips in RISC-V-based flexible health monitoring wearables.

Finally, it is noteworthy that the execution time of our optimization framework (Section~\ref{sec:optimization}) is negligible, averaging 20 minutes and ranging from 5 to 30 minutes.
\section{Conclusion}

In this work, we propose and implement compact and energy-efficient ML-accelerated flexible RISC-V microprocessors, targeting conformable healthcare wearables.
Our bespoke MAC-based co-processor for the SERV core, maximizes performance and energy efficiency.
Our proposed constrained programming-based optimization automatically selects co-processor constants and decomposes MLP operations on the available hardware, enabling a tailored microprocessor for each model and aligning rapid fabrication with equally fast design cycles.
Post-layout results across healthcare datasets demonstrate near-real-time inference within $2.42$mm$^2$, under the power budget of flexible batteries, and requiring on average $4200$ gates, enabling realistic implementations despite the inherent limitations of flexible electronics.

\section{Acknowledgment}
This work is partially supported by the European Research Council (ERC) (Grant No. 101052764) and co-funded by the H.F.R.I call “Basic Research Financing (Horizontal support of all Sciences)” under the National Recovery and Resilience Plan “Greece 2.0” (H.F.R.I. Project Number: 17048).

\bibliographystyle{IEEEtran}
\bibliography{IEEEabrv,references}

\end{document}